\pdfoutput=1

\documentclass[11pt]{article}

\usepackage[]{emnlp2021}

\usepackage{times}
\usepackage{latexsym}

\usepackage[T1]{fontenc}

\usepackage[utf8]{inputenc}

\usepackage{microtype}

\usepackage{multirow}
\usepackage{graphicx}
\usepackage{algorithm}
\usepackage[noend]{algpseudocode}
\usepackage{enumitem}
\usepackage{booktabs}

\title{Data Augmentation for Low-Resource Named Entity Recognition Using Backtranslation}

\newcommand*{\affaddr}[1]{#1} 
\newcommand*{\affmark}[1][*]{\textsuperscript{#1}}

\author{Usama Yaseen\affmark[1,2], Stefan Langer\affmark[1,2]\\ 
 \affaddr{\affmark[1]Technology, Siemens AG  Munich, Germany}\\
  \affaddr{\affmark[2]CIS, University of Munich (LMU) Munich, Germany} \\
  {\tt \{usama.yaseen,langer.stefan\}@siemens.com}
}

\begin{document}
\maketitle
\begin{abstract}

The state of art natural language processing systems relies on sizable training datasets to achieve high performance. Lack of such datasets in the specialized low resource domains lead to suboptimal performance. In this work, we adapt backtranslation to generate high quality and linguistically diverse synthetic data for low-resource named entity recognition. We perform experiments on two datasets from the materials science (MaSciP) and biomedical domains (S800). The empirical results demonstrate the effectiveness of our proposed augmentation strategy, particularly in the low-resource scenario.

\end{abstract}

\section{Introduction}

Most recently, various deep learning methods have demonstrated state of the art performance for many natural language processing tasks such as text classification, sentiment analysis and named entity recognition. The availability of large training datasets is crucial to achieve this improved performance and avoid overfitting. However, in many real-world applications collecting such large training data is not possible. This is especially true for specialized domains, such as the material science or biomedical domain, where annotating data requires expert knowledge and is usually time-consuming and expensive.

Data augmentation (DA) \cite{SimardLDV96} has been investigated to overcome this low resource problem. Label preserving synthetic data generation is widely used in computer vision \cite{KrizhevskySH12, CiresanMS12, FawziSTF16} and speech domains \cite{SchluterG15, KoPPSK17}. The discrete nature of language makes it difficult to adapt data augmentation strategies from computer vision and speech to natural language processing. Unlike computer vision, where hardcoded transformations (such as rotation, masking, cropping etc.) can be easily applied without changing the label semantics, the manipulation of a single word in a sentence could change its meaning.

Recently, there is an increased interest in applying data augmentation to natural language processing tasks. Most augmentation methods explore sentence-level tasks such as sentiment analysis \cite{LiestingFT21},  text classification \cite{WeiZ19, Xie19} and sentence-pair tasks such as natural language inference \cite{MinMDPL20} and machine translation \cite{WangPDN18}. The augmentation methods either employ heuristics such as word replacement \cite{ZhangZL15, WangPDN18, CaiCSZZY20}, word swap \cite{SahinS18, MinMDPL20} or random deletion \cite{WeiZ19}  to generate augmented instances by manipulating a few words in the original sentence; or generate completely artificial instances via sampling from generative models such as variational autoencoders \cite{YooSL19, MesbahYSTLBH19} or backtranslation models \cite{YuDLZ00L18, IyyerWGZ18}.

The sequence labelling tasks such as named entity recognition (NER) and part-of-speech tagging (POS) involves prediction at the token level. This makes applying token-level transformation difficult as such manipulations may change the corresponding token level label. The existing DA methods for sequence labelling uses dependency tree morphing \cite{SahinS18}, MIXUP \cite{ZhangCDL18} to generate queried samples in the active learning scenario \cite{ZhangYZ20}, sample novel sequences from a trained language model \cite{ding-etal-2020-daga} and apply pre-defined heuristics such as label-wise token and synonym replacement \cite{DaiA20}. The existing sequence labelling DA methods are limiting as they: a). rely on linguistics resources like dependency parser or WordNet b). involves training a language model c). generate grammatically incoherent sequences d). cannot generate linguistically diverse sequences.

Motivated by the advancements in machine translation and the availability of high-quality machine translation systems \cite{hytra-He15, WuSCLNMKCGMKSJL16, Junczys-Dowmunt19}, in this work we adapt backtranslation to the task of NER. Backtranslation (BT) can automatically generate diverse paraphrases of a sentence or a phrase by naturally injecting linguistic variations. The injected linguistic variations can be further diversified by introducing layers of intermediate language translations. In this work, we generate paraphrases of one or several phrases in a sentence. We empirically demonstrate the effectiveness of our proposed method on two domain-specific NER datasets.

\section{Related Work}

There is an abundance of recent work on DA methods for NLP tasks, we refer the readers to \citeauthor{Feng-Gamgal21} for an extensive survey. In this section we narrow our focus to existing DA methods for sequence labelling tasks like NER and POS. We categorize existing DA methods for sequence labelling into two categories:

{\bf Rule-based:} DA primitives, which use predefined easy-to-compute transformations. We briefly describe six of such transformations proposed in the existing work:

\begin{enumerate}[label=(\alph*)]
  \item {\it NER::Label-wise token replacement (LwTR):} Replace a token with another token of the same entity type at random \cite{DaiA20}.
  \item {\it NER::Synonym replacement (SR):} Replace a token with one of its synonyms retrieved from WordNet at random \cite{DaiA20}.
  \item {\it NER::Mention replacement (MR):} Replace an entity mention with another entity mention of the same entity type at random \cite{DaiA20}.
  \item {\it NER::Shuffle within segments (SiS):} Divide the sequence of tokens into segments of the same label and then randomly shuffle the order of segments \cite{DaiA20}.
  \item {\it POS::Crop Sentences:} Given a dependency tree of the sentence, "crop" a sentence by removing dependency links \cite{SahinS18}.
    \item {\it POS::Rotate Sentences:} Given a dependency tree of the sentence, "rotate" a sentence by moving the tree fragments around the root \cite{SahinS18}.
\end{enumerate}

{\bf Generative models:} The existing work uses pre-trained language models to generate either part of the sequence or the entire sequence with the corresponding NER tags. \citeauthor{Min-Kye21} proposed {\it Filtered BERT} which randomly masks one or several tokens in the original sentence and let BERT \cite{DevlinCLT19} predict the masked token. The augmentation is only accepted if the cosine similarity of the word embeddings (computed using fastText embeddings \cite{BojanowskiGJM17}) of the original token and the predicted masked token is above a certain threshold. \citeauthor{ding-etal-2020-daga} propose a two-step DA process DAGA. First, a shallow language model is trained over linearized sequences of tags and words. Second, sequences are sampled from this language model and delinearized to create new examples.

\section{Data Augmentation via Backtranslation}

\begin{figure}[t]
    \centering
    \includegraphics[scale=0.40]{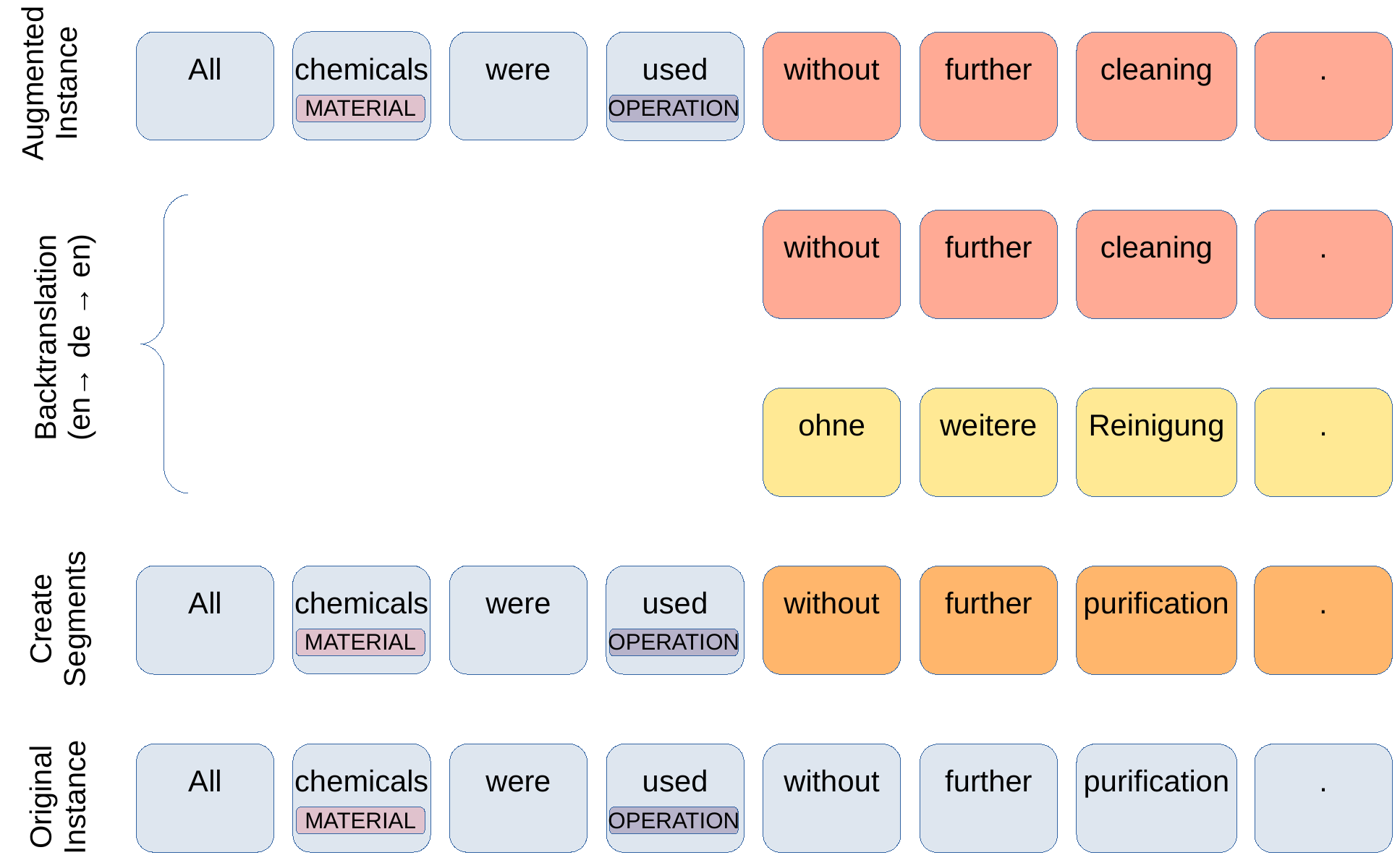}
    \caption{An illustration of data augmentation via \emph{backtranslation} for NER. Note that backtranslation is only applied to the context around the entity mentions. Here the entity mention context is first translated to German and then back to English using an off-the-shelf machine translation system. The backtranslation results in a paraphrase of the original entity mention context. The original entity mention context is replaced with backtranslated context to create the augmented data instance.}
    \label{fig:backtranslation-ner}
\end{figure}

Figure \ref{fig:backtranslation-ner} illustrates an example of data augmentation using \emph{backtranslation} for NER with German as an intermediate language. 
In a nutshell, the algorithm consists of three steps. First, the input token sequence is split into segments of the same label; thus, each segment corresponds to either the entity mention or the context around the entity mention. Note that only context around the entity mention is a candidate for the backtranslation. Second, the validity of the segment is determined based on the length of the segment, we only consider segments with three or more tokens as a valid segment for backtranslation. As a final step, the segment tokens are translated to the intermediate language(s) and finally back to the source language; the original segment tokens are replaced with the backtranslated tokens and thus we obtain the augmentation of the original input token sequence. In practice, we use a binomial distribution to randomly decide whether the segment should be backtranslated. Since only the context around the entity mention is backtranslated, it is straightforward to adjust the corresponding BIO-label sequence accordingly for the backtranslated text.

Data augmentation with backtranslation augments the original training set with diverse paraphrases of the entity mention contexts to help the underlying NER model to generalize beyond the standard training set.

\section{Experiments and Results}
\subsection{Datasets}

\begin{table*}[t]
\begin{small}
    \setlength{\tabcolsep}{2pt}
    \centering
    \begin{tabular}{r | r | cccc | r | cccc | r}
    \toprule
    \multirow{2}{*}{\bf Embeddings} & \multirow{2}{*}{\bf Method} & \multicolumn{4}{c|}{\bf MaSciP} & \multirow{2}{*}{\bf $\Delta$} & \multicolumn{4}{c|}{\bf S800} & \multirow{2}{*}{\bf $\Delta$} \\
    \cline{3-6}
    \cline{8-11}
    & & S & M & L & F & & S & M & L & F & \\ \midrule
    \multirow{7}{*}{Glove} & None & 48.52\scriptsize{$\pm$ 3.5} & 67.98\scriptsize{$\pm$ 0.5} & 73.02\scriptsize{$\pm$ 0.8} & 75.37\scriptsize{$\pm$ 0.3} &  & 12.24\scriptsize{$\pm$ 1.6} & 21.61\scriptsize{$\pm$ 0.7} & 49.99\scriptsize{$\pm$ 2.6} & 60.44\scriptsize{$\pm$ 1.4} & \\  
    & LwTR & 61.95\scriptsize{$\pm$ 1.3} & 68.04\scriptsize{$\pm$ 0.7} & 75.05\scriptsize{$\pm$ 0.3} & 75.32\scriptsize{$\pm$ 0.2} & 2.9 & 17.37\scriptsize{$\pm$ 0.4} & 41.19\scriptsize{$\pm$ 1.3} & 50.93\scriptsize{$\pm$ 1.8} & 62.46\scriptsize{$\pm$ 1.2} & 6.9 \\  
    & SR & {\bf 63.91\scriptsize{$\pm$ 1.6}} & 69.44\scriptsize{$\pm$ 0.7} & 75.10\scriptsize{$\pm$ 0.4} & 76.95\scriptsize{$\pm$ 0.8} & 4.6 & 17.83\scriptsize{$\pm$ 1.3} & 43.86\scriptsize{$\pm$ 1.1} & 57.76\scriptsize{$\pm$ 0.2} & 65.28\scriptsize{$\pm$ 0.5} & 10.1 \\  
    & MR & 63.46\scriptsize{$\pm$ 0.3} & 69.64\scriptsize{$\pm$ 0.7} & 75.08\scriptsize{$\pm$ 0.4} & 76.33\scriptsize{$\pm$ 1.0} & 4.6 & 17.86\scriptsize{$\pm$ 2.4} & 43.90\scriptsize{$\pm$ 0.8} & 56.70\scriptsize{$\pm$ 0.9} & 65.34\scriptsize{$\pm$ 0.6} & 9.9 \\  
    & SiS & 63.63\scriptsize{$\pm$ 1.1} & 69.60\scriptsize{$\pm$ 0.3} & 73.35\scriptsize{$\pm$ 0.2} & {\bf 77.36\scriptsize{$\pm$ 0.3}} & 4.6 & 17.17\scriptsize{$\pm$ 1.7} & 44.36\scriptsize{$\pm$ 0.2} & 56.80\scriptsize{$\pm$ 0.9} & 64.93\scriptsize{$\pm$ 0.2} & 9.7 \\  
     & BT & 63.66\scriptsize{$\pm$ 0.6} & {\bf 69.67\scriptsize{$\pm$ 0.1}} & {\bf 75.22\scriptsize{$\pm$ 0.2}} & 76.85\scriptsize{$\pm$ 0.4} & {\bf 4.6} & {\bf 31.06\scriptsize{$\pm$ 1.7}} & {\bf 47.82\scriptsize{$\pm$ 1.2}} & {\bf 58.86\scriptsize{$\pm$ 1.0}} & {\bf 66.89\scriptsize{$\pm$ 0.3}} & {\bf 15.1} \\   
    \midrule
    \multirow{7}{*}{SciBERT} & None & 61.89\scriptsize{$\pm$ 1.3} & 71.76\scriptsize{$\pm$ 0.6} & 78.52\scriptsize{$\pm$ 0.1} & 79.91\scriptsize{$\pm$ 0.1} & & 39.78\scriptsize{$\pm$ 1.6} & 51.15\scriptsize{$\pm$ 1.6} & 64.08\scriptsize{$\pm$ 0.8} & 72.73\scriptsize{$\pm$ 0.9} & \\ 
    & LwTR & 66.88\scriptsize{$\pm$ 1.4} & 73.40\scriptsize{$\pm$ 1.1} & 77.83\scriptsize{$\pm$ 0.1} & 77.51\scriptsize{$\pm$ 3.0} & 0.9 & 41.37\scriptsize{$\pm$ 0.4} & 51.76\scriptsize{$\pm$ 1.0} & 64.97\scriptsize{$\pm$ 1.6} & 71.34\scriptsize{$\pm$ 0.1} & 0.4 \\  
    & SR & 67.07\scriptsize{$\pm$ 0.8} & 74.56\scriptsize{$\pm$ 0.3} & 78.47\scriptsize{$\pm$ 0.4} & 79.71\scriptsize{$\pm$ 0.3} & 1.9 & 40.24\scriptsize{$\pm$ 1.2} & {\bf 53.68\scriptsize{$\pm$ 0.4}} & 62.98\scriptsize{$\pm$ 1.4} & 71.77\scriptsize{$\pm$ 0.6} & 0.2 \\  
    & MR & 67.65\scriptsize{$\pm$ 1.0} & 74.60\scriptsize{$\pm$ 1.3} & 78.04\scriptsize{$\pm$ 1.1} & 79.57\scriptsize{$\pm$ 0.6} & 1.9 & 41.89\scriptsize{$\pm$ 1.4} & 53.24\scriptsize{$\pm$ 1.3} & 66.56\scriptsize{$\pm$ 1.2} & 70.87\scriptsize{$\pm$ 0.5} & 1.2 \\  
    & SiS & 66.87\scriptsize{$\pm$ 2.9} & 73.40\scriptsize{$\pm$ 1.5} & {\bf 78.95\scriptsize{$\pm$ 0.6}} & 79.79\scriptsize{$\pm$ 0.5} & 1.7 & 41.57\scriptsize{$\pm$ 1.8} & 51.83\scriptsize{$\pm$ 0.7} & 65.16\scriptsize{$\pm$ 1.0} & 71.20\scriptsize{$\pm$ 0.6} & 0.5 \\  
     & BT & {\bf 70.11\scriptsize{$\pm$ 0.8}} & {\bf 75.86\scriptsize{$\pm$ 0.8}} & 78.92\scriptsize{$\pm$ 0.2} & {\bf 80.30\scriptsize{$\pm$ 0.5}} & {\bf 3.3} & {\bf 44.60\scriptsize{$\pm$ 1.0}} & 53.22\scriptsize{$\pm$ 1.3} & {\bf 66.76\scriptsize{$\pm$ 1.1}} & {\bf 72.92\scriptsize{$\pm$ 0.2}} & {\bf 2.4} \\ 
    \bottomrule
    \end{tabular}
    \caption{F1-score on test sets using different subsets of the training set. Here: {\bf S}, {\bf M}, {\bf L} and {\bf F} refer to {\it small} (50 instances), {\it medium} (150 instances), {\it large} (500 instances) and {\it full} (all instances) set. We repeat all experiments three times with different seeds. Mean values and standard deviations are reported. $\Delta$ column shows the averaged improvement due to data augmentation for each embedding type across the datasets.}
    \label{table-results}
\end{small}
\end{table*}

We empirically evaluate backtranslation for NER on two English datasets from the materials science and biomedical domains: MaSciP \cite{MysoreJKHCSFMO19}\footnote{\url{https://github.com/olivettigroup/annotated-materials-syntheses}} and S800 \cite{Pafilis2013TheSA}\footnote{\url{https://github.com/spyysalo/s800}}. MaSciP contains synthesis procedures annotated with synthesis operations and their typed arguments. S800 consists of PubMed abstracts annotated for organism mentions. We use the original train-dev-test split provided by the authors. The descriptive statistics of the datasets are reported in Appendix (see Table \ref{table-data-statistics}).

We simulate low-resource setting as proposed by \citeauthor{DaiA20}; we select 50, 150, 500 sentences from the training set to create the corresponding small, medium and large training sets (denoted as S, M, L in Table \ref{table-results}, whereas the complete training set is denoted as F). Data augmentation is only applied on the training set without altering the development and test set.

\subsection{NER Model}

\begin{figure}[t]
    \centering
    \includegraphics[scale=0.45]{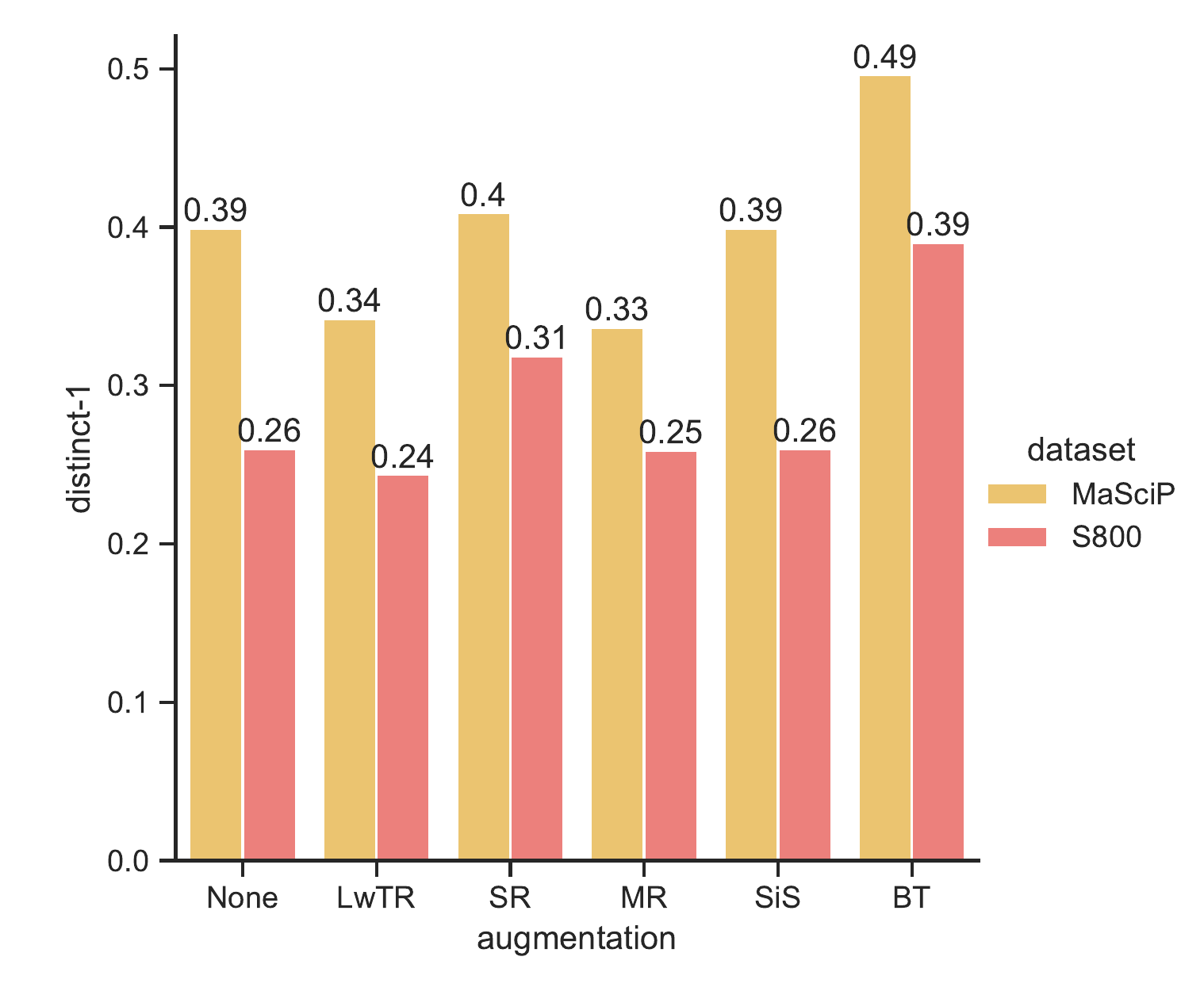}
    \caption{The diversity statistics of various augmentation techniques across the datasets.}
    \label{fig:diversity-statistics}
\end{figure}

We follow the standard approach of modelling the NER task as a sequence labelling task. The mainstream sequence labelling models for NER employ the neural-based encoder and an output tagging component. The typical choice of the encoder is a sequence model such as LSTM \cite{HochreiterS97} or more recently a sequence encoder such as Transformer \cite{VaswaniSPUJGKP17}; the output tagging component is usually a conditional random field layer \cite{LaffertyMP01} to model dependencies between neighbouring labels. 

We employed the standard \emph{BiLSTM-CRF} model \cite{LampleBSKD16} as our backbone model. We experimented with context-independent \emph{GloVe} embeddings \cite{PenningtonSM14} as well as state-of-the-art contextualized \emph{BERT} embeddings \cite{DevlinCLT19}. We employed \emph{SciBERT} \cite{BeltagyLC19}, which is based on the BERT model pretrained on scientific publications; our preliminary experiments suggest that SciBERT achieves better performance than BERT. The superiority of domain-specific BERT models on downstream tasks has been observed by existing studies \cite{GururanganMSLBD20, DaiA20}.

We report the micro-average F\textsubscript{1} score as an evaluation metric. We employ early stopping and report the F\textsubscript{1} score on the test set using the best performant model on the development set.

\subsection{Backtranslation Models}

We employed the Huggingface’s Transformers library \cite{wolf-etal-2020-transformers} port of the pretrained English$\leftrightarrow$German models \cite{NgYBOAE19}\footnote{\url{https://huggingface.co/facebook/wmt19-en-de}}\textsuperscript{,}\footnote{\url{https://huggingface.co/facebook/wmt19-de-en}} as the underlying backtranslation models for all our experiments.

\subsection{Hyperparameters}

Following existing work \cite{DaiA20}, we tune the number of augmentation instances per training instance from a list of numbers: $\{1, 3, 6, 10\}$. When all data augmentation methods are applied, this tuning list is reduced to: $\{1, 2, 3\}$. We also tune the probability value $p$ of the beta distribution which is used to decide if the segment in a sequence should be backtranslated. It is searched over a list of numbers: $\{0.1, 0.3, 0.5, 0.7\}$. We perform a grid search over these two hyperparameters to find their best combination on the development set.

\subsection{Results}

We report the performance of various augmentation techniques on the test sets in Table \ref{table-results}. For the most part, all data augmentation techniques improve over the baseline; backtranslation results in the biggest average improvement for both context-independent {\it GloVe} and contextualized {\it SciBERT} embeddings under different data usage percentiles. We attribute the improved performance of backtranslation to the generation of linguistically diverse and meaning-preserving {\it entity mention contexts} to enable better generalization of the underlying NER model.


The data augmentation techniques contribute to the biggest improvement in performance when the training sets are small, this effect is reduced as the training sets get larger (see columns {\bf S}  vs {\bf F}   in Table \ref{table-results}).  The augmentation on the complete training set even decreases the performances for some augmentation techniques. The performance impact of data augmentation on varying sizes of training sets has also been observed in the existing work \cite{FadaeeBM17a, DaiA20, ding-etal-2020-daga}. 

We also investigate the effectiveness of data augmentation techniques on the mainstream contextualized ({pretrained \it SciBERT}) embeddings. All the augmentation techniques especially backtranslation result in better performance when compared to the baseline. However, the average performance improvement due to data augmentation with SciBERT embeddings is lower as compared to the GloVe embeddings. 

To quantitatively measure the diversity introduced by various augmentation techniques, we report {\it distinct-1} \cite{LiGBGD16} in Figure \ref{fig:diversity-statistics}. Distinct-1 quantifies the intra-text diversity based on distinct unigrams in each sentence, the value is scaled by the total number of tokens in the sentence to avoid favouring long sentences. Backtranslation yield the highest level of unigram diversity, this is not very surprising as backtranslation is known to generate diverse linguistic variations. 

\section{Conclusion}

In this paper, we adapt backtranslation to the token-level sequence tagging task of NER. We show that backtranslation can generate high-quality coherent and linguistically diverse synthetic data for NER. The experiments on two domain-specific datasets demonstrate the effectiveness of backtranslation as a competitive data augmentation strategy for NER.


\bibliography{custom}
\bibliographystyle{acl_natbib}

\clearpage
\appendix

\section{Datasets}
\label{sec:datasets}

\begin{table}[h]
	\center
	\small
	\renewcommand*{\arraystretch}{1.25}
	\resizebox{.50\textwidth}{!}{
	\setlength\tabcolsep{3.pt}
	\begin{tabular}{r | ccc | ccc  }
		& \multicolumn{3}{c|}{\bf MaSciP} & \multicolumn{3}{c}{\bf S800} \\
		\cline{2-7}
		& Train & Dev & Test & Train & Dev & Test \\
		Number of sentences & 1,899 & 112 & 162 & 5,733 &830 & 1,630 \\ 
		Number of mentions & 18,896 & 1,190 & 1,259 & 2,557 & 384 & 767 \\ 
		Number of unique mentions  & 4,707 & 590 & 605 & 1,070 & 194 & 3781 \\ 
		Number of entity types & 21 & 20 & 21 & 1 & 1 & 1 \\
	\end{tabular}
}
\caption{The descriptive statistics of the datasets.}
\label{table-data-statistics}
\end{table}

\end{document}